\documentclass[11pt]{article}
\usepackage{amssymb}

\usepackage[final]{acl}

\usepackage{times}
\usepackage{latexsym}
\usepackage{hyperref}

\usepackage[T1]{fontenc}
\usepackage[T5]{fontenc}

\usepackage[utf8]{inputenc}

\usepackage{float}
\usepackage{microtype}

\usepackage{inconsolata}
\usepackage{subcaption}

\usepackage{amsmath}
\usepackage{graphicx}
\usepackage{multirow}

\usepackage{xcolor, color, soul}

\sethlcolor{yellow}

\usepackage[flushleft]{threeparttable}

\title{MTikGuard System: A Transformer-Based Multimodal System for Child-Safe Content Moderation on TikTok}

\author{
Dat Thanh Nguyen\textsuperscript{1,4,5},
Nguyen Hung Lam\textsuperscript{3,4,5},
Anh Thi-Hoang Nguyen\textsuperscript{1,4,5},
Trong-Hop Do\textsuperscript{2,4,5}
\\[1ex]
\textsuperscript{1} Faculty of Information Science and Engineering, University of Information Technology\\
\textsuperscript{2} Faculty of Software Engineering, University of Information Technology\\
\textsuperscript{3} Faculty of Computer Science, University of Information Technology \\
\textsuperscript{4} University of Information Technology, Ho Chi Minh City, Vietnam \\
\textsuperscript{5}Vietnam National University, Ho Chi Minh City, Vietnam \\
\texttt{Emails: \{22520224,22520968\}@gm.uit.edu.vn, \{anhnth,hopdt\}@uit.edu.vn}
}

\begin{document}
\maketitle
\begin{abstract}
With the rapid rise of short-form videos, TikTok has become one of the most influential platforms among children and teenagers, but also a source of harmful content that can affect their perception and behavior. Such content, often subtle or deceptive, challenges traditional moderation methods due to the massive volume and real-time nature of uploads. This paper presents \textbf{MTikGuard}, a real-time multimodal harmful content detection system for TikTok, with three key contributions: (1) an \emph{extended} TikHarm dataset expanded to 4,723 labeled videos by adding diverse real-world samples, (2) a multimodal classification framework integrating visual, audio, and textual features to achieve \textbf{state-of-the-art} performance with 89.37\% accuracy and 89.45\% F1-score, and (3) a scalable streaming architecture built on Apache Kafka and Apache Spark for real-time deployment. The results demonstrate the effectiveness of combining dataset expansion, advanced multimodal fusion, and robust deployment for practical large-scale social media content moderation. The dataset is available at \url{https://github.com/ntdat-8324/MTikGuard-System.git}. 


\textbf{Disclaimer:} This paper contains images and content from social networks that may be considered sensitive, harmful, or related to adult content and suicide.
\end{abstract}

\section{Introduction}
Harmful video content refers to videos containing visuals, audio, or messages that may negatively impact viewers’ cognition, emotions, and behaviors - particularly those of children. Such content may involve depictions of violence, explicit language, antisocial behavior, hate speech, or dangerous challenges disguised as “fun” activities that children can easily imitate. In the era of short-form video as a dominant mode of entertainment, the spread of harmful content occurs faster than ever before, often beyond the supervision of parents or schools. Children, with limited analytical and self-protective capabilities, are especially vulnerable to these negative role models, often failing to recognize their long-term consequences for personality development and behavior. Consequently, the early detection and timely warning of harmful video content are essential to maintaining a safe online environment for young audiences.

Online platforms, particularly social media, enable harmful videos to proliferate rapidly, making control and prevention increasingly challenging. To address this issue, we build upon and extend the publicly available TikHarm dataset to improve model training effectiveness and develop a system that can curb the spread of harmful trends targeting children.

In this study, we expand the TikHarm dataset by collecting additional videos following the original selection criteria, thereby improving the model’s learning capacity and robustness in real-world deployment. Leveraging this extended dataset, we introduce \textbf{MTikGuard}, \textit{a real-time harmful content detection system for TikTok}. MTikGuard processes multiple modalities from videos, including visual frames and textual information (extracted via OCR and speech-to-text), enabling richer contextual understanding (see Appendix~\ref{sec:task_definition} for the formal task definition). With a multimodal transformer-based architecture, MTikGuard outperforms single-modal baselines, particularly in detecting harmful trends presented in subtle or disguised forms. Our main contributions are summarized as follows:
\begin{itemize}
    \item We expand the TikHarm dataset by collecting and annotating an additional 775 videos, resulting in a total of 4,723 labeled samples. This extension improves model robustness and generalization.
    \item We propose a transformer-based multimodal architecture for harmful video detection that integrates visual frames, OCR-extracted text, and speech-to-text transcripts. By jointly modeling these modalities, the system captures nuanced cross-modal cues, enabling more accurate identification of harmful content, even when disguised or indirectly presented.
    \item We introduce the MTikGuard system, a real-time video analysis and detection pipeline for harmful content. Beyond its robust multimodal model, MTikGuard integrates advanced technologies such as Apache Spark, Apache Kafka, Docker, and Apache Airflow, enabling scalable, reliable deployment for detecting harmful content from online sources in production environments.
\end{itemize}

\section{Related Work}
Many studies have focused on detecting inappropriate content using only video data. \cite{10.1145/3297280.3297487} introduced the KidsGUARD system, which utilizes an LSTM-based autoencoder to learn video representations from features extracted by the VGG16 CNN network, aiming to detect unsafe content for children. The model achieved a recall of 81\% and a precision of 80\% on a dataset of 109,835 video clips. \cite{9696242} proposed a method using EfficientNet-B7 for feature extraction and a BiLSTM network for video representation to classify inappropriate content in animated YouTube videos, achieving 95.66\% accuracy and an F1-score of 92.67\%. \cite{article} conducted a study applying computer vision techniques to detect e-cigarette-related content in TikTok videos, building an object detection model based on YOLOv7 that achieved a recall of 77\%, precision of 86.30\%, and F1-score of 81.40\%. Transformer-based models such as TimeSformer ~\cite{bertasius2021timesformer}, VideoMAE, and ViViT have also been effectively applied in video classification, enhancing accuracy and efficiency in detecting inappropriate content. Notably, TikGuard \cite{10753192} employs advanced Transformer models like TimeSformer, VideoMAE, and ViViT to classify TikTok content, achieving 86.7\% accuracy on the TikHarm dataset, demonstrating the potential of these models in detecting harmful content for children.

Recognizing the limitations of relying solely on video data, recent studies have shifted towards multimodal approaches, combining information from video, audio, and text to improve the detection of inappropriate content. \cite{phukan2025snifrboostingfinegrained} proposed SNIFR, a new framework that leverages Transformers to integrate audio and visual information for detecting harmful content targeting children, outperforming single-modality methods. \cite{das2023hatemmmultimodaldatasethate} introduced HateMM, a multimodal dataset comprising approximately 43 hours of video from BitChute labeled as hate or non-hate, along with frame-level rationale annotations for label decisions. Their findings showed that incorporating all modalities significantly improved overall hate speech detection performance (accuracy = 79.80\%, macro F1-score = 79\%), with a gain of approximately 5.7\% in macro F1-score compared to the best single-modality model.

\section{Dataset}

\subsection{TikHarm Dataset}
TikHarm is a dataset developed to support the task of detecting harmful content on the TikTok platform, with a particular focus on protecting children from such content. The TikHarm dataset is available at \href{https://www.kaggle.com/datasets/anhoangvo/tikharm-dataset/data}{Link} . The dataset consists of 3,948 videos collected from TikTok through unofficial APIs, focusing on popular hashtags and keywords commonly appearing in child-oriented content. Videos were selected based on engagement metrics (views, likes, comments) while ensuring diversity in topics and posting times.

After the collection process, all videos were manually annotated into four content categories: \textbf{Safe}, \textbf{Adult Content}, \textbf{Harmful Content}, and \textbf{Suicide}. The annotation was carried out by a team of trained experts, following unified guidelines to minimize subjective bias. The inter-annotator agreement, measured by the Fleiss’ Kappa coefficient, reached 81.25\%, indicating high consistency and reliability in the classification process.

The dataset is split into three subsets: \textit{training}, \textit{development}, and \textit{testing}. Table~\ref{distribution-TikHarm} presents the number of samples, average duration, and total video duration for each split.

\begin{table}[h]
  \centering
  \resizebox{0.9\linewidth}{!}{%
  \begin{tabular}{lcccc}
    \hline
    & \multicolumn{4}{c}{\textbf{Duration (s)}} \\
    \cline{2-5}
    & \textbf{Samples} & \textbf{Min} & \textbf{Max} & \textbf{Avg} \\
    \hline
    Train set     & 2,762  & 3.88 & 600 & 38.71 \\
    Dev. set      & 396    & 1.95 & 600 & 38.77 \\
    Test set      & 790    & 5.04 & 600 & 38.57 \\
    \hline
  \end{tabular}
  }
  \caption{\label{distribution-TikHarm}
    Distribution of Video Samples and Duration by Data Split in the TikHarm Dataset.
  }
\end{table}

The class distribution of the entire dataset is detailed in Table~\ref{class-distribution-TikHarm}. It can be observed that videos in the \textbf{Safe} class tend to have a significantly longer average duration, reflecting the characteristics of educational or wholesome entertainment content. Conversely, the \textbf{Suicide} class shows a much shorter average duration, partly due to TikTok’s strict moderation policies.  
With balanced class distribution and reliable annotations, TikHarm serves as a robust foundation for training and evaluating deep learning models in harmful content detection tasks. Notably, it is the first dataset in Vietnam dedicated to TikTok content safety for children, supporting research in video processing and real-time content moderation systems.

\begin{table}[h]
  \centering
  \resizebox{\linewidth}{!}{%
  \begin{tabular}{lcccc}
    \hline
    & \multicolumn{4}{c}{\textbf{Duration (s)}} \\
    \cline{2-5}
    & \textbf{Samples} & \textbf{Min} & \textbf{Max} & \textbf{Avg} \\
    \hline
    Safe                 & 997  & 16.96 & 65.36 & 18.10 \\
    Adult Content        & 977  & 9.84  & 36.25 & 18.10 \\
    Harmful Content      & 990  & 9.88  & 35.92 & 18.10 \\
    Suicide              & 984  & 4.63  & 16.96 & 18.10 \\
    \hline
  \end{tabular}
  }
  \caption{\label{class-distribution-TikHarm}
    Distribution of Video Samples and Duration by Label in the TikHarm Dataset.
  }
\end{table}

\subsection{Extended Dataset}
\label{sec:extended_dataset}

One of the \textbf{main contributions} of this study is the construction and release of an Extended Dataset (available at \url{https://www.kaggle.com/datasets/kusnguyen/extra-dataset}) to enhance the coverage and representation of trending harmful content, thereby improving the detection of more subtle and sophisticated forms of harmful material on TikTok. This dataset was collected and annotated using a rigorous procedure derived from the original TikHarm dataset, ensuring consistent quality and labeling criteria, while incorporating new samples that reflect current content trends on the platform.

\paragraph{Dataset Collection.}
The supplementary data collection began by visiting TikTok’s hashtag pages to identify trending hashtags. These hashtags were categorized based on their potential to yield videos falling into the four labels of the dataset. From this filtered list, we randomly collected 1,000 videos using Selenium in combination with the TikTok-Content-Scraper tool. The selection was guided by engagement metrics (views, likes, comments) and ensured diversity in topics and posting times, consistent with the original TikHarm collection methodology.

\paragraph{Dataset Annotation.}
\begin{figure}[t]
	\centering
	\includegraphics[width=\columnwidth]{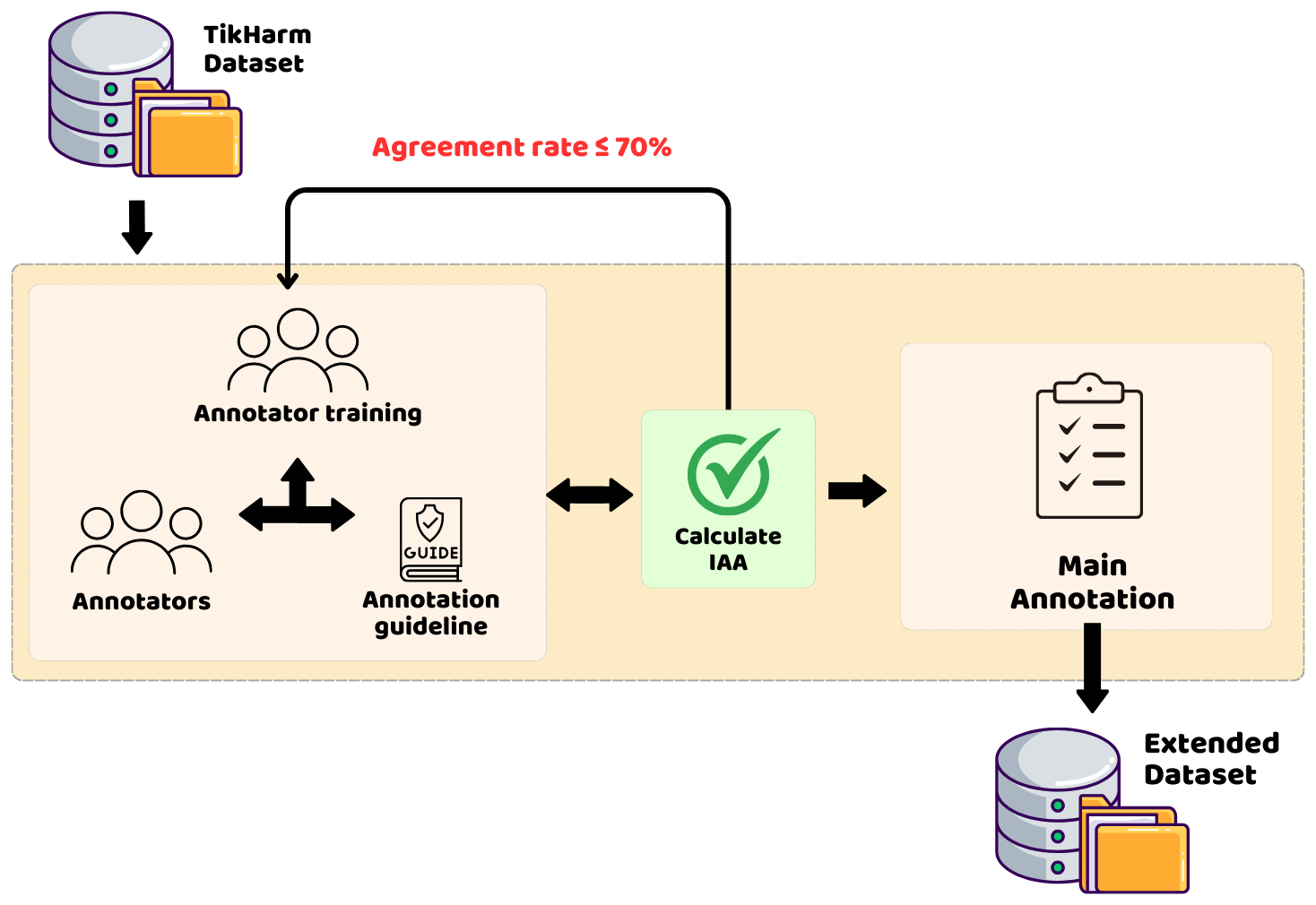}
	\caption{The dataset extending process.}
	\label{fig:annotate_process}
\end{figure}
Figure~\ref{fig:annotate_process} illustrates the overall dataset extending process. Before annotation, we refined the labeling guidelines by clearly defining each label and providing detailed illustrative examples. This ensured the extended dataset maintained high quality and alignment with the original TikHarm annotations.  
Three annotators, thoroughly trained on the updated guidelines, first labeled a test set of 100 samples from the original dataset. Their agreement with the ground truth achieved Cohen’s Kappa scores of 0.82, 0.78, and 0.80, reflecting substantial to almost perfect agreement.  
Inter-annotator analysis revealed very high consensus, with two annotators achieving a Cohen’s Kappa of 0.97. The overall Fleiss’ Kappa across all three annotators reached 0.86, indicating almost perfect agreement. These results validated the annotators’ consistent application of the labeling criteria, enabling each annotator to proceed independently with the remaining extended data. 

After annotation and removal of low-quality samples, we obtained 775 qualified videos. Unlike the original TikHarm dataset, which was intentionally balanced across labels, the extended dataset has an imbalanced label distribution (Table~\ref{class-distribution-extended-dataset}). This imbalance was intentionally preserved to better reflect the real-world distribution of harmful content on TikTok, where certain categories appear more frequently than others. Such a distribution allows models to learn from “in-the-wild” data, improving their generalization in real-world scenarios. Further details of the annotation guidelines are presented in Appendix~\ref{sec:annotation_guidelines}.

\begin{table}[h]
  \centering
  \resizebox{\linewidth}{!}{%
  \begin{tabular}{lcccc}
    \hline
    & \multicolumn{4}{c}{\textbf{Duration (s)}} \\
    \cline{2-5}
    & \textbf{Samples} & \textbf{Min} & \textbf{Max} & \textbf{Avg} \\
    \hline
    Safe                 & 251  & 3.04 & 60 & 25.97 \\
    Adult Content        & 164  & 4.69 & 60 & 18.78 \\
    Harmful Content      & 203  & 5.06 & 60 & 22.61 \\
    Suicide              & 157  & 6.00 & 60 & 19.83 \\
    \hline
  \end{tabular}
  }
  \caption{\label{class-distribution-extended-dataset}
    Distribution of Video Samples and Duration by Label in the Extended Dataset.
  }
\end{table}

\subsection{Final Combined Dataset}
The Extended Dataset (Section~\ref{sec:extended_dataset}) was merged with TikHarm to form the Final Combined Dataset, which integrates both the balanced distribution of the original dataset and the realistic distribution of the extended one. This combination provides a unique benchmark that not only retains comparability with prior studies (by keeping the original TikHarm test set intact) but also challenges models with a more representative and diverse range of harmful content.  

The extended samples were split into training and development sets at an 8.5:1.5 ratio, while the TikHarm test set was kept unchanged to ensure fair evaluation against existing methods. The detailed distribution of samples across the three datasets is presented in Table~\ref{samples-distribution-3-dataset}.

\begin{table}[h]
  \centering
  \resizebox{\linewidth}{!}{%
  \begin{tabular}{lcccc}
    \hline
    \textbf{Dataset} & \textbf{Train} & \textbf{Dev.} & \textbf{Test} & \textbf{Total} \\
    \hline
    TikHarm                 & 2,762  & 396 & 790 & 3,948 \\
    Extended Dataset        & 656    & 119 & -   & 775 \\
    Final Combined Dataset  & 3,418  & 515 & 790 & 4,723 \\
    \hline
  \end{tabular}
  }
  \caption{\label{samples-distribution-3-dataset}
    Sample distribution across TikHarm, Extended Dataset, and the Final Combined Dataset.
  }
\end{table}

By uniting the curated balance of TikHarm with the real-world skew of the Extended Dataset, the Final Combined Dataset represents one of the most comprehensive and practical resources for harmful content detection on TikTok to date, making it a key contribution of this research.

\section{Method}
\label{sec:method}

\subsection{Data Processing}

Data preprocessing is a crucial step to ensure high-quality input for the multimodal classification system, especially in the noisy and diverse context of short-form TikTok videos. We design a processing pipeline that extracts and refines linguistic information from both the audio and the visual frames of each video.

\paragraph{Text Extraction from Audio.}
The audio signal is first separated from the video using the \texttt{librosa} library, with each sample limited to a maximum duration of 60 seconds to optimize processing efficiency. After normalization, low-energy segments are removed based on the Root Mean Square (RMS) criterion. Speech recognition is performed in parallel for Vietnamese and English using the \textit{Whisper large-v3} model \cite{radford2022whisper}. The hypothesis with the most complete contextual coverage is retained. To reduce hallucinated or generic outputs, we apply a filter to remove short or repetitive sentences, as well as phrases without meaningful content (e.g., ``thank you'', ``hello'', ``bye''). In addition, a specialized spam filter, implemented via regular expressions, is used to remove advertising text, engagement calls, or meaningless repetitions common in social-media content.

\paragraph{Text Extraction from Frames.}
In parallel with the audio processing, text appearing in the video frames is extracted using \textit{EASYOCR}\footnote{https://github.com/JaidedAI/EasyOCR}, which supports Vietnamese and English. To balance recognition accuracy and computational cost, two representative frames are selected at 30\% and 70\% of the total video duration. Each frame is resized so that its longer side does not exceed 640 pixels. The OCR results are then processed through the same spam and language filters as the audio transcripts. Only Vietnamese or English text that is non-duplicated and passes the noise filter is retained.

\paragraph{Synthesis and Standardization of Text Input.}
The retained transcripts from both modalities (audio and OCR) are concatenated into a single string following the template:
\texttt{Audio: ... | OCR: ...}.  
A lightweight language detector \cite{joulin2016bagtricksefficienttext} ensures that only Vietnamese or English content is preserved, while all other languages and non-linguistic text are removed. This process produces a compact, semantically rich textual input that reduces noise and improves the reliability of downstream classification.

\subsection{Multimodal Harmful Content Detection Architecture}

\begin{figure*}[t]
	\centering
	\includegraphics[width=0.9\linewidth]{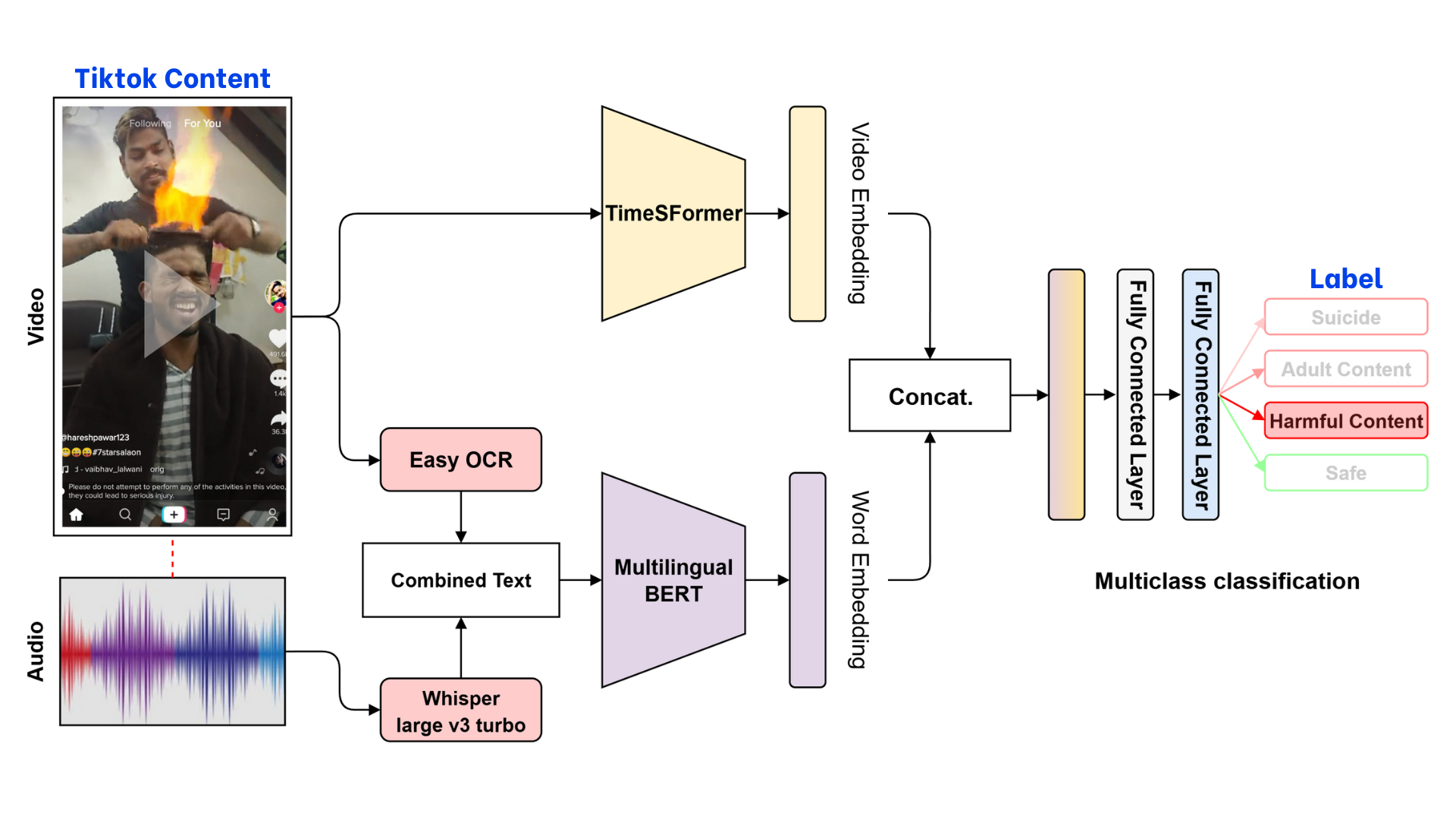}
	\caption{Proposed multimodal harmful content detection architecture.}
	\label{classification_enhancement_pipeline}
\end{figure*}

The proposed architecture (Figure~\ref{classification_enhancement_pipeline}) leverages the strengths of two pretrained encoders: \textit{TimeSFormer} for video representation and a transformer-based language model for text representation.

\paragraph{Video Encoder.}
We adopt \textit{TimeSFormer}, an advanced transformer-based model that applies the Divided Space--Time Attention mechanism to separately model spatial and temporal dependencies. This design has been shown to improve classification efficiency and achieve state-of-the-art performance on large-scale benchmarks such as Kinetics-400 and Kinetics-600.

\paragraph{Text Encoder.}
Since the content of the videos primarily includes Vietnamese and English, we use \textit{Multilingual DistilBERT} \cite{sanh2019distilbert} to effectively capture multilingual context. We also experiment with \textit{ViSoBERT} \cite{nguyen2023visobert}, a language model optimized for Vietnamese social-media text. Each encoder output is passed through a separate projection block consisting of a linear layer, a ReLU activation, and dropout to reduce overfitting.

\paragraph{Feature Fusion and Classification.}
The projected features from the video and text encoders are concatenated and passed through a fusion network composed of multiple linear layers, Layer Normalization, ReLU activations, and dropout. The resulting joint multimodal representation is then fed into a final linear layer to perform four-way classification into \textit{Safe}, \textit{Adult Content}, \textit{Harmful Content}, and \textit{Suicide}. This late-fusion strategy preserves the strengths of each individual modality while enabling the system to exploit cross-modal cues, improving its generalization ability in multimodal harmful content detection. In addition to the late-fusion strategy, we also design an architecture with attention-based fusion. Specifically, similar to prior studies on multimodal learning, after concatenating the two modalities, the combined features are passed through Multihead Self-Attention layers, which allow the model to capture global relationships between textual and visual features, enrich the joint representation, and focus on the most relevant information.

\section{Experiments}
\label{sec:experiments}

\subsection{Experimental Configurations}
We conducted a series of offline experiments to evaluate the effectiveness of the proposed multimodal architecture for harmful video content detection. The evaluation was performed on both the original \textit{TikHarm} dataset and an \textit{expanded dataset}, where the latter was constructed by augmenting \textit{TikHarm} with additional crawled samples in a train–validation ratio of 8.5:1.5.

For the video encoder, we adopted the pretrained \textit{TimeSFormer} model, while for the text encoder, we experimented with \textit{Multilingual DistilBERT} and \textit{ViSoBERT} to determine the most suitable language model for integration into the actual system.

To ensure fairness across experiments, we fixed the hyperparameters for all runs: training was conducted for 6 epochs with a batch size of 8 for both training and evaluation. The initial learning rate was set to 1e-4, combined with a warmup ratio and weight decay, and gradients were accumulated every two steps. Losses on the training and validation sets were computed every 100 steps, and performance was evaluated using accuracy, macro-precision, macro-recall, and macro-F1. The macro formulation was chosen to mitigate class imbalance by giving equal weight to each class, including minority labels, thereby reducing bias toward dominant classes - a common challenge in harmful content classification tasks.

We retained up to three checkpoints per training run and selected the best model based on the highest macro-F1 score on the validation set. The final reported results on the test set were obtained using the checkpoint with the best macro-F1 score. The implementation was carried out in \textit{PyTorch} and executed on the Kaggle platform with an Intel(R) Xeon(R) CPU @ 2.00GHz and an NVIDIA Tesla P100 GPU (16 GB, CUDA 11.4).

\subsection{Main Results}
\begin{table*}[h]
  \centering
  \begin{tabular}{llcccc}
    \hline
    \multirow{2}{*}{\textbf{Dataset}} & \multirow{2}{*}{\textbf{Language Model}} & \multicolumn{2}{c}{\textbf{Validation set}} & \multicolumn{2}{c}{\textbf{Test set}} \\
    \cline{3-6}
    & & \textbf{Acc.} & \textbf{F1} & \textbf{Acc.} & \textbf{F1} \\
    \hline
    \multirow{5}{*}{TikHarm} 
    & TimeSFormer-only & 86.66 & 86.62 & 86.71 & 86.68 \\
    \cline{2-6}
    & Multilingual BERT & 88.64 & 88.56 & \textbf{89.37} & \textbf{89.45} \\
    \cline{2-6}
    & Multilingual BERT [$\blacklozenge$] & 88.13 & 88.14 & \underline{88.35} & \underline{88.38} \\
    \cline{2-6}
    & ViSoBERT & \underline{89.14} & \underline{89.19} & 87.22 & 87.18 \\
    \cline{2-6}
    & ViSoBERT [$\blacklozenge$] & \textbf{90.15} & \textbf{90.17} & 87.97 & 87.87 \\
    \hline
    \multirow{4}{*}{Final Combined}
    & Multilingual BERT & \textbf{80.78} & \textbf{80.77} & \underline{88.35} & \underline{88.31} \\
    \cline{2-6}
    & Multilingual BERT [$\blacklozenge$] & \underline{80.58} & 80.32 & \textbf{88.48} & \textbf{88.44} \\
    \cline{2-6}
    & ViSoBERT & \underline{80.58} & \underline{80.41} & 87.47 & 87.39 \\
    \cline{2-6}
    & ViSoBERT [$\blacklozenge$] & 80.39 & 80.22 & 88.10 & 88.03 \\
    \hline
  \end{tabular}
  \caption{\label{tab:experimental_results}
    Performance Comparison of Multimodal Models on TikHarm and Final Combined Datasets. The table shows classification results for various language models integrated with TimeSFormer, evaluated on the Validation and Test sets. The notation [$\blacklozenge$] denotes models employing attention-based fusion to integrate visual and textual features. Bold values represent the best result, while underlined values signify the second-best performance within each dataset group. The TimeSFormer-only row refers to the baseline architecture utilizing only the visual encoder without any language model, with results adopted directly from \cite{10753192}.
  }
\end{table*}


As shown in Table~\ref{tab:experimental_results}, on the original TikHarm dataset, the results demonstrate the clear advantage of multimodal integration with attention-based fusion. On the validation set, ViSoBERT with attention-based fusion achieved the highest performance, surpassing its concat-based late-fusion variant. This indicates that when linguistic and visual features are deliberately integrated at the attention layer, the model can better exploit cross-modal interactions to reinforce classification decisions. On the TikHarm test set, Multilingual BERT achieved the highest macro-F1 score of 89.45 (Acc 89.37), followed by its attention-based variant. The inversion of rankings between the validation and test sets suggests that the cross-lingual generalization capability of Multilingual BERT is advantageous when evaluating on unseen data. In contrast, TimeSFormer-only, relying solely on visual features, performed substantially worse (F1 86.62 on validation and 86.68 on test), reinforcing the critical role of textual information in decoding the semantics of short-form videos.

On the Final Combined dataset (our extended version), validation performance dropped slightly to around 80 F1, with Multilingual BERT achieving the highest score (F1 80.77; Acc 80.78), followed by ViSoBERT (F1 80.41; Acc 80.58). This decline is expected, as the dataset was deliberately expanded to validate model robustness under real-world conditions with label imbalance and ambiguous cases. On the test set, however, models maintained strong performance close to 88\% F1. Multilingual BERT with attention-based fusion achieved the best result (F1 88.44; Acc 88.48), slightly outperforming its concat late-fusion counterpart (F1 88.31; Acc 88.35) and both ViSoBERT configurations. This finding suggests that when data becomes more diverse and noisy, attention-based fusion enables the model to better capture cross-modal interactions, thereby improving the detection and classification of harmful video content in real-world scenarios.

Notably, since both the original and extended datasets naturally include noisy audio, blurred frames, and code-switched language, the strong performance across all settings indicates that MTikGuard remains robust under such imperfect conditions. This highlights the system’s ability to generalize well even in noisy, real-world TikTok scenarios, with qualitative examples provided in Appendix~\ref{sec:noisy_cases}.

\subsection{Ablation Study}

To better understand the contribution of each modality, we conducted an ablation study where the model was trained and evaluated using only one modality at a time: \textbf{OCR}, \textbf{ASR}, or \textbf{Video}. The results on both TikHarm and the Final Combined Dataset are presented in Table~\ref{tab:ablation_results}.


\begin{table*}[h]
  \centering
  \begin{tabular}{lccccccc}
    \hline
    \multirow{2}{*}{\textbf{Dataset}} & \multicolumn{3}{c}{\textbf{Modality}} & \multicolumn{2}{c}{\textbf{Validation set}} & \multicolumn{2}{c}{\textbf{Test set}} \\
    \cline{2-8}
    & \textbf{Video} & \textbf{OCR} & \textbf{ASR} & \textbf{Acc.} & \textbf{F1} & \textbf{Acc.} & \textbf{F1} \\
    \hline
    \multirow{4}{*}{TikHarm} 
    & $\times$ &   &   & \textbf{88.64} & \textbf{88.64} & \underline{88.99} & \underline{88.92} \\
    \cline{2-8}
    &   & $\times$ &   & 36.11 & 31.33 & 35.44 & 30.27 \\
    \cline{2-8}
    &   &   & $\times$ & 46.72 & 47.05 & 49.11 & 48.88 \\
    \cline{2-8}
    & $\times$ & $\times$ & $\times$ & \textbf{88.64} & \underline{88.56} & \textbf{89.37} & \textbf{89.45} \\
    \hline
    \multirow{4}{*}{Final Combined}
    & $\times$ &   &   & \underline{80.58} & \underline{80.44} & \underline{87.97} & \underline{87.90} \\
    \cline{2-8}
    &   & $\times$ &   & 33.40 & 27.77 & 36.2 & 30.52 \\
    \cline{2-8}
    &   &   & $\times$ & 44.85 & 44.72 & 50.51 & 50.26 \\
    \cline{2-8}
    & $\times$ & $\times$ & $\times$ & \textbf{80.78} & \textbf{80.77} & \textbf{88.35} & \textbf{88.31} \\
    \hline
  \end{tabular}
  \caption{\label{tab:ablation_results}
    Ablation Study on Modality Effectiveness Using the Best-Performing Architecture.
    This table shows the contribution of each modality: Video, OCR and ASR, using the best model configuration on TikHarm Dataset (last row) as the baseline. The $\times$ symbol indicates the modality was retained; removing it (empty cell) demonstrates its individual impact on performance (Acc. and F1).
  }
\end{table*}

The results indicate that the \textbf{Video} modality plays the most critical role, achieving nearly 89\% accuracy and F1-score on the TikHarm dataset, and maintaining high performance on the Final Combined Dataset. 
\textbf{ASR} contributes moderately, with F1 around 47–50\%, showing that speech transcripts provide useful but less stable information due to noise, informal speech, or lack of context. 
\textbf{OCR} performs weakest, below 37\% accuracy and around 30\% F1, largely because textual content in TikTok videos is often sparse, noisy, or unrelated (e.g., ads or captions).
Overall, although the Video modality dominates, combining it with ASR and OCR in a multimodal fusion setup significantly improves performance, particularly for subtle or disguised harmful content that is difficult to detect with visuals alone.

\subsection{Discussion}

\begin{figure}[h]
	\centering
	\includegraphics[width=.45\textwidth]{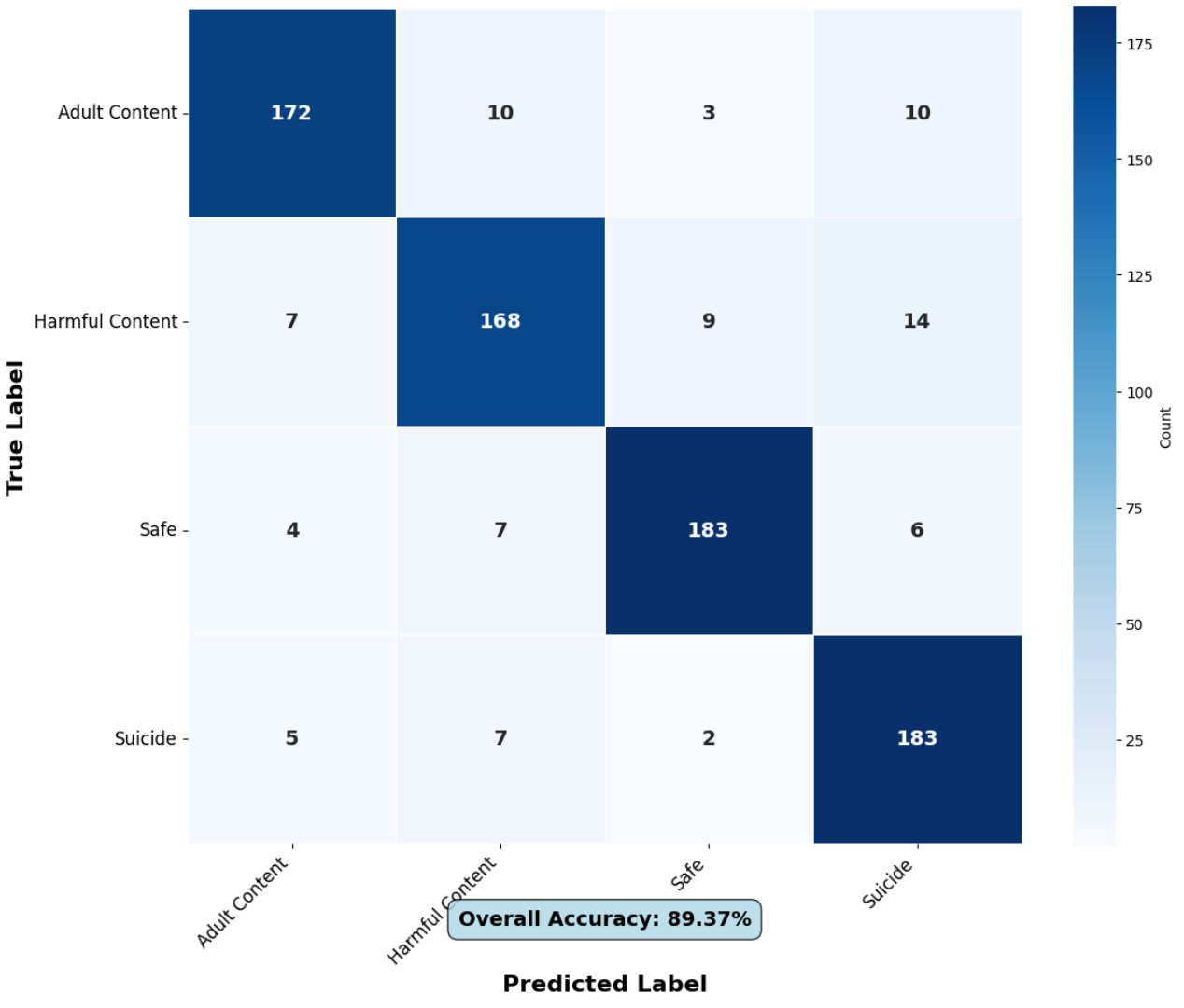}
	\caption{Confusion matrix of the best-performing model on the test set.}
	\label{fig:confusion_matrix}
\end{figure}

The confusion matrix (Figure~3) reveals that the system achieved an overall accuracy of \textbf{89.37\%} on the test set. The \textit{Safe} category was most accurately predicted (183 correct classifications) with minimal mislabeling, indicating high reliability in detecting non-harmful content. \textit{Suicide} also performed strongly (183 correct predictions) but showed some confusion with \textit{Adult Content} and \textit{Harmful Content}. While \textit{Adult Content} achieved 172 correct identifications, the \textit{Harmful Content} category (168 correct predictions) exhibited the \textbf{highest misclassification rate into \textit{Suicide}}, clearly suggesting overlapping expression patterns in sensitive themes.

We adopted a multi-class approach due to the observed content overlap and the implicit hierarchy among harmful labels (e.g., Suicide is more severe than Harmful). Therefore, the authors of the TikHarm dataset and we prioritized the maximum risk in the labeling process, assigning a single, definitive label to each piece of content. This choice simplifies the decision-making process into a clear, automated action (BLOCK/ALLOW), avoiding the complexity of multi-label output. To manage cases where the model has low confidence (is "confused"), we will utilize Label Smoothing during training and set a confidence threshold to flag ambiguous cases for human review, thus effectively managing the overlap without altering the core label structure.

Besides the intrinsic issue of label overlap, data imbalance is also a challenging factor, particularly affecting the minority class, \textit{Suicide}. We observed higher misclassification between \textit{Suicide} and \textit{Harmful/Adult content}. Nevertheless, we retained this imbalance to reflect real-world TikTok distributions, where suicide-related videos are rarer and often subtle. This improves generalization in deployment, though future work may adopt reweighting strategies or focal loss to further boost minority-class recall.

TikTok videos often contain noisy audio, blurred frames, and code-switched language, which naturally introduce errors in OCR and ASR. Despite these challenges, MTikGuard demonstrates strong robustness thanks to multimodal fusion, where errors in one modality can be compensated by signals from another. For instance, background noise may reduce ASR quality, but OCR-extracted harmful text can still support correct classification, and vice versa. Ablation results (Table~\ref{tab:ablation_results}) confirm this effect: unimodal OCR or ASR alone performs poorly, while multimodal fusion consistently improves accuracy by about 2–3\%. Moreover, Multilingual DistilBERT effectively handles code-switching (e.g., Vietnamese speech with English captions). Together, these findings highlight the system’s reliability in noisy, real-world TikTok scenarios.

Overall, the model demonstrated robust discriminative ability across all categories despite the diverse and unpredictable nature of TikTok short videos.

\section{MTikGuard System}
\label{sec:mtikguard}

\begin{figure*}[t]
    \centering
    \includegraphics[width=0.9\linewidth]{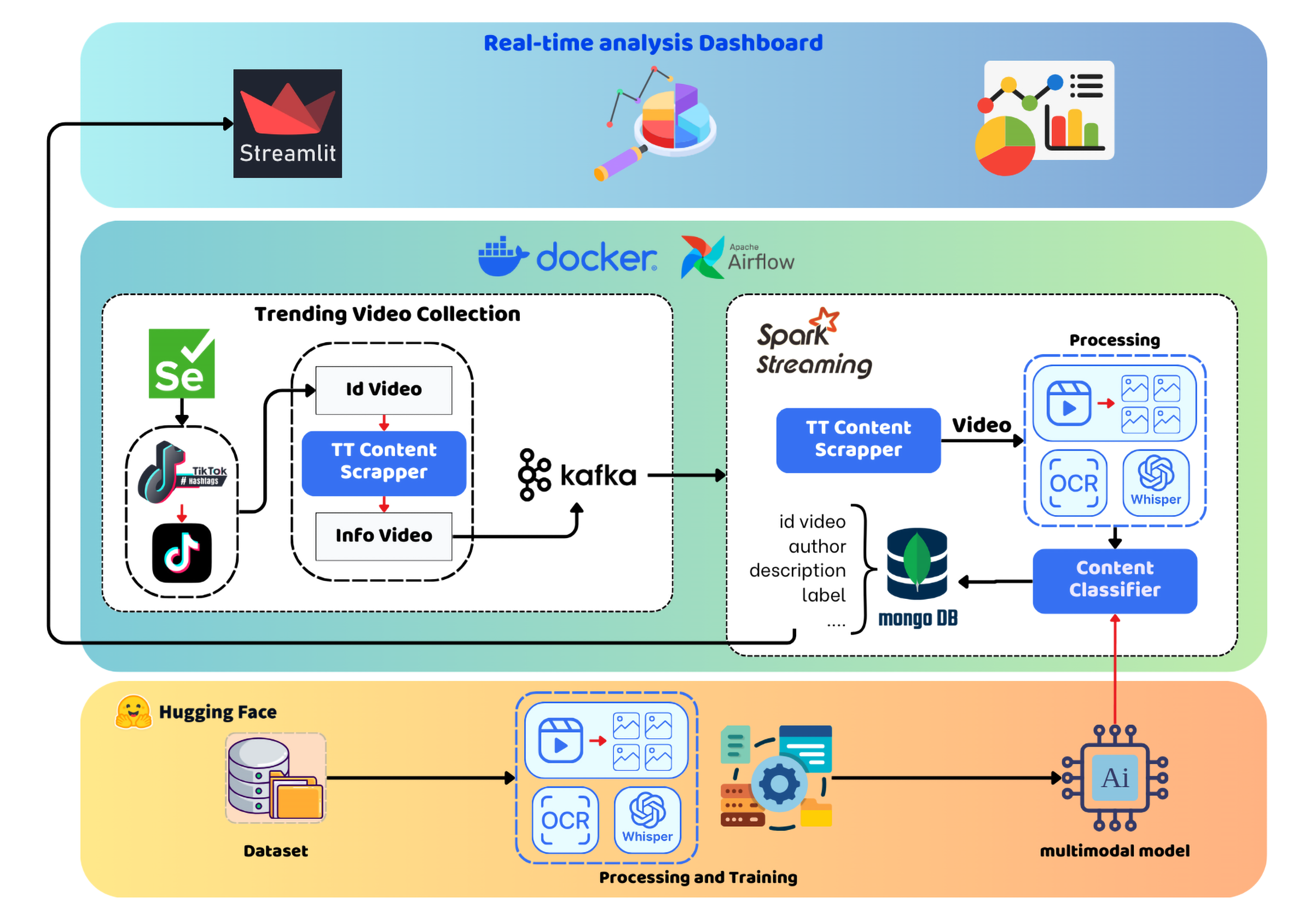}
    \caption{Overview of the MTikGuard system.}
    \label{fig:mtikguard_system}
\end{figure*}

Figure~\ref{fig:mtikguard_system} illustrates the overall architecture of \textbf{MTikGuard}, a modular and extensible system designed for real-time multimodal content moderation on TikTok. The design ensures adaptability to various deployment environments and scalability to handle high-throughput streaming data. The system is structured into three primary functional layers, presented from bottom to top as follows:

The foundation layer focuses on dataset preparation and model training. The system leverages the TikHarm dataset, augmented with an extended dataset to enhance coverage and robustness. For each video, visual features are extracted using the TimeSFormer video transformer, while textual features are derived from transcribed audio via Whisper and scene text via EasyOCR. Text representations are encoded using either Multilingual DistilBERT or ViSoBERT, as described in previous sections. The resulting multimodal architecture is trained and evaluated using macro-F1 as the primary selection metric. Once the optimal model is obtained, it is serialized and integrated into the real-time censorship pipeline.

The middle layer implements the automated streaming pipeline, orchestrated with Apache Airflow\footnote{https://airflow.apache.org} and containerized using Docker\footnote{https://www.docker.com} for portability and reproducibility. The pipeline includes:
\begin{itemize}
    \item A Selenium-based crawler\footnote{https://www.selenium.dev} together with the TikTok Content Scraper module retrieves trending TikTok videos based on hashtags, extracts metadata and video IDs, and publishes them to an Apache Kafka\footnote{https://kafka.apache.org} topic .
    \item Kafka consumers download the videos, extract text from audio using Whispe and from frames using EasyOCR, and then classify the content using the trained multimodal model. The classification results, along with metadata, are stored in MongoDB Atlas \footnote{https://www.mongodb.com} for further querying and analysis.
    \item Airflow coordinates the execution of producers and consumers, monitors pipeline health, and manages periodic reporting, ensuring stability and scalability.
\end{itemize}

The top layer emphasizes the modular nature of MTikGuard, allowing flexible deployment in both cloud-based and on-premise environments. Each component - such as Kafka consumers, inference services, or data storage - can be scaled independently. The architecture also facilitates integration with additional modalities or extension to other short-video platforms in future work.

\section{Conclusion}

We introduced \textbf{MTikGuard}, a real-time multimodal detection system for harmful TikTok content that jointly analyzes visual, audio, and textual modalities. By extending the TikHarm dataset to 4,723 annotated videos and leveraging transformer-based encoders (TimeSFormer and DistilBERT) within a scalable streaming pipeline, MTikGuard achieves \textbf{state-of-the-art performance} with 89.37\% accuracy and 89.45\% macro-F1. These results highlight the effectiveness of combining dataset expansion, multimodal fusion, and practical deployment infrastructure in safeguarding young users from harmful online content. In addition, our pipeline is inherently multilingual: Whisper and EasyOCR support dozens of languages, while Multilingual DistilBERT enables robust cross-lingual text encoding. Thus, extending to other regions would require minimal adjustments, primarily re-training with a small annotated subset to adapt to local linguistic nuances.

While the current system adopts a late-fusion strategy, it does not yet explore more advanced mechanisms to dynamically weight each modality’s contribution. In future work, we plan to enhance audio processing to better handle non-speech cues and incorporate attention-driven multimodal fusion to improve robustness and interpretability in complex real-world scenarios.

\section*{Acknowledgments}
This research was supported by The VNUHCM-University of
Information Technology's Scientific Research Support Fund.

\newpage
\bibliography{custom}

\appendix
\section{Appendix}
\subsection{Task Definition}
\label{sec:task_definition}

The harmful content detection task in TikTok videos is formulated as a supervised \textbf{multimodal classification} problem. Given a video $V$, the system processes three modalities:  
(1) \textit{visual frames} $F_v$,  
(2) \textit{audio transcripts} $T_a$ (from speech-to-text), and  
(3) \textit{scene text} $T_o$ (from OCR on frames).  
After noise removal and normalization, each modality is encoded into feature vectors: $x_v$ for visual data and $x_t$ for text (combining $T_a$ and $T_o$). These are fused using a late fusion strategy:
\begin{equation}
    x_m = \text{Fusion}(x_v, x_t)
\end{equation}
The classifier $f_{\theta}$ maps $x_m$ to one of four labels:
\begin{equation}
\resizebox{0.89\linewidth}{!}{%
$y \in \{\text{Safe},\ \text{Adult Content},\ \text{Harmful Content},\ \text{Suicide}\}$
}
\end{equation}

The goal is to minimize cross-entropy loss on the labeled dataset $\mathcal{D}$ while maximizing the \textbf{macro-F1 score} to ensure balanced performance across classes.

\begin{figure}[h]
	\centering
	\includegraphics[width=0.9\columnwidth]{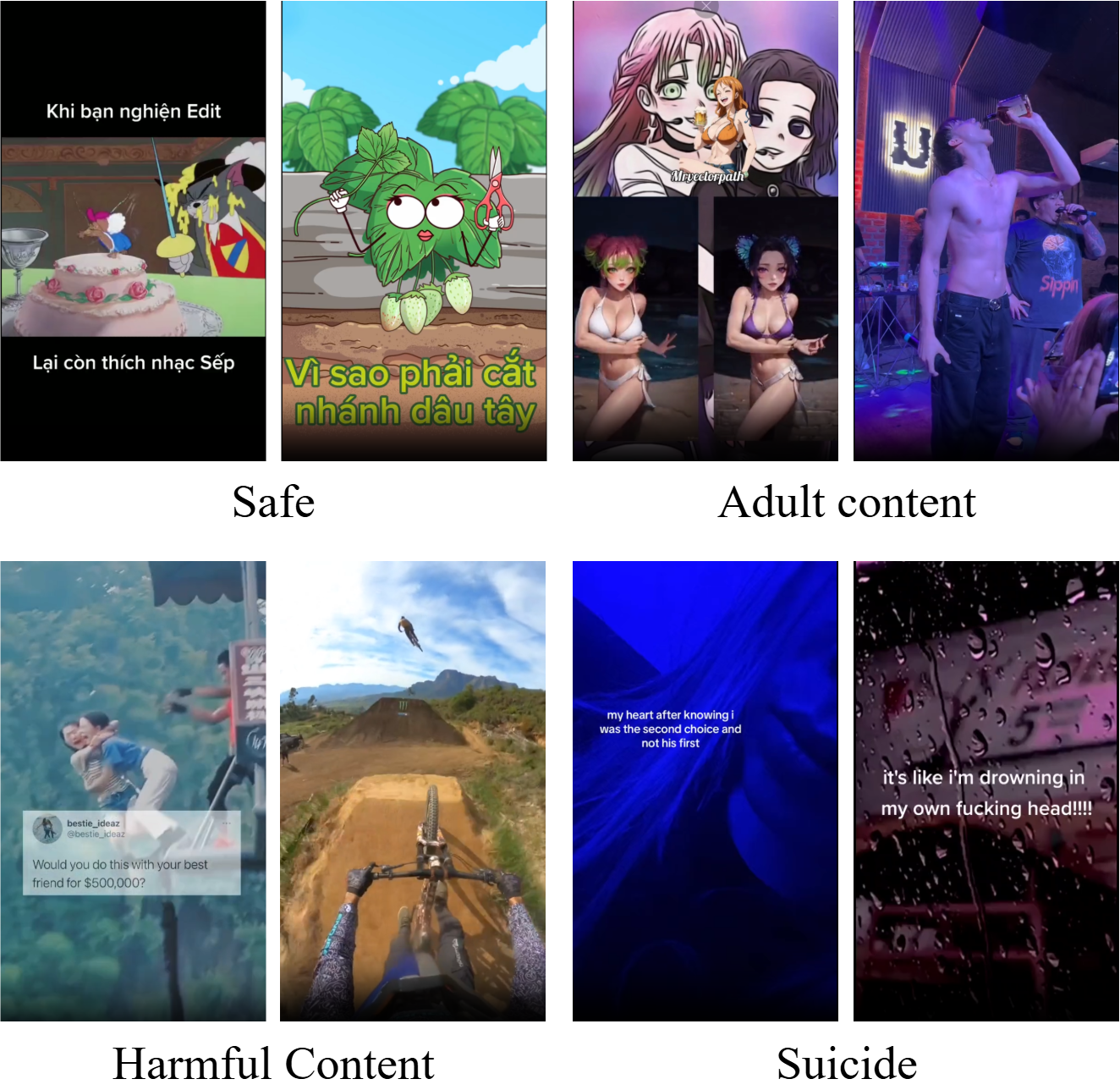}
	\caption{An example of the Harmful Content Detection  task.}
	\label{fig:example}
\end{figure}

\subsection{Annotation Guidelines}
\label{sec:annotation_guidelines}
\subsubsection{Annotation Tool}

\begin{figure}[h]
	\centering
	\includegraphics[width=0.9\columnwidth]{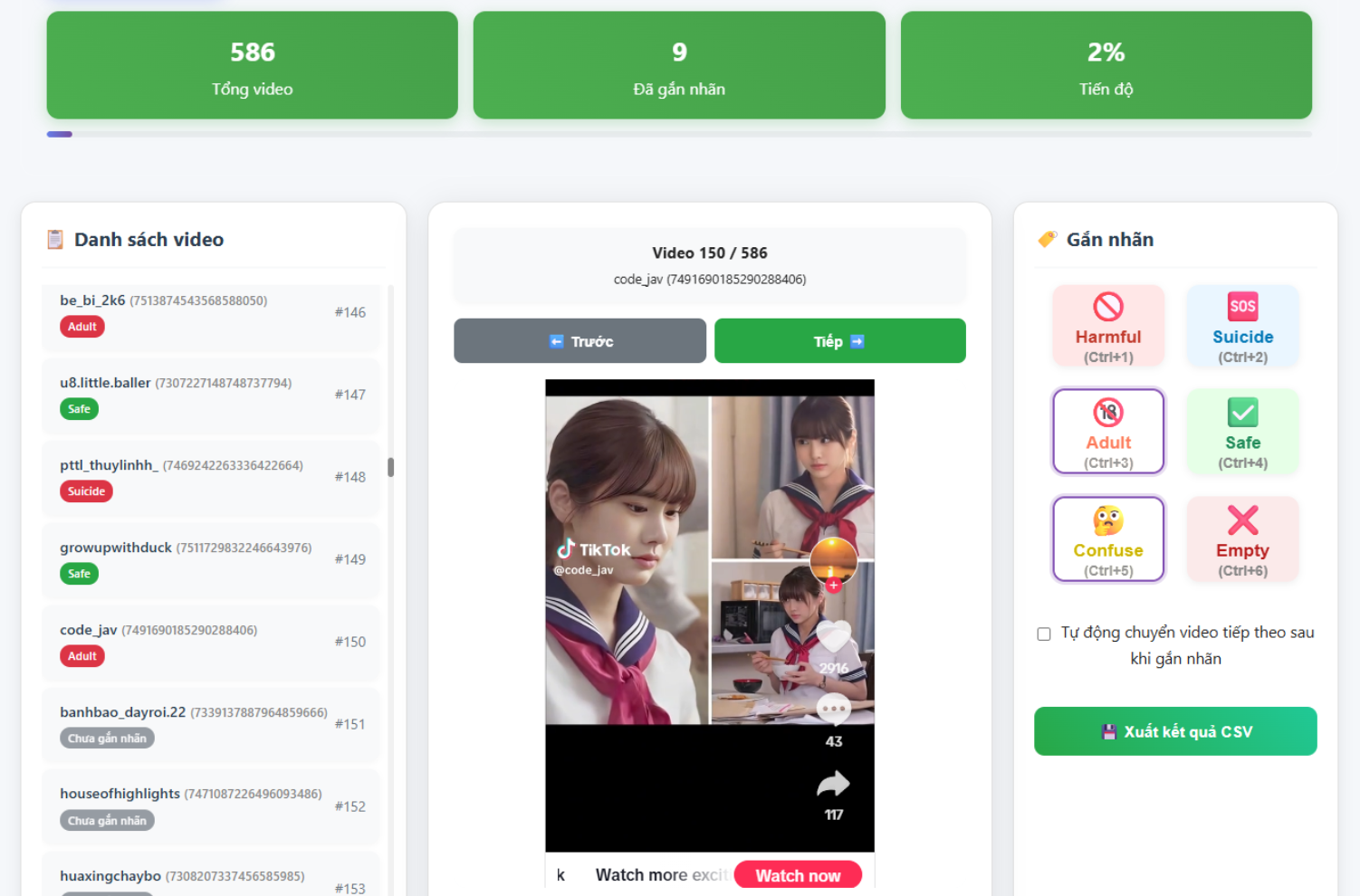}
	\caption{The web application interface for annotators.}
	\label{fig:annotate_tool}
\end{figure}

To support this complex annotation task, we developed a dedicated web-based annotation tool (available at \url{https://github.com/Kussssssss/tiktok-labeling-tool.git}).  The system displays a list of videos and associated metadata (user name and video ID), a video player for observing content directly, and a label selection panel. The interface also provides progress tracking, together with two auxiliary options, \textit{confuse} and \textit{empty}, to handle ambiguous cases and ensure labeling quality and consistency (Figure~\ref{fig:annotate_tool}).

\subsubsection{Annotation Guidelines}

Each annotator must strictly follow the label definitions derived from the TikHarm dataset. Only one label can be assigned to each video; multiple labels are considered invalid. Samples that are unclear are first labeled based on the annotator’s best judgment and later cross-validated with others.

\begin{itemize}
    \item \textbf{Safe:} Content that is appropriate and safe for children, without violence, sexuality, strong language, or dangerous behaviour. \textit{(e.g., educational programs, family-friendly vlogs, cartoons, DIY activities)}.
    \item \textbf{Adult Content:} Content containing sexual elements, nudity, or language not suitable for children, including sexually suggestive material. \textit{(e.g., provocative dancing, lingerie try-ons, sexual jokes or innuendo)}.
    \item \textbf{Harmful Content:} Content depicting violence, dangerous acts, or harmful behaviour that may negatively influence children. \textit{(e.g., fights, bullying, dangerous challenges involving weapons or chemicals)}.
    \item \textbf{Suicide:} Content referring to suicidal behaviour, intentions, or methods that may negatively impact viewers, especially minors. \textit{(e.g., self-harm attempts, suicide tutorials, confessions of suicidal thoughts)}.
\end{itemize}

\subsection{Error Analysis Details}

\begin{figure}[h]
\centering
\includegraphics[width=0.143\textwidth]{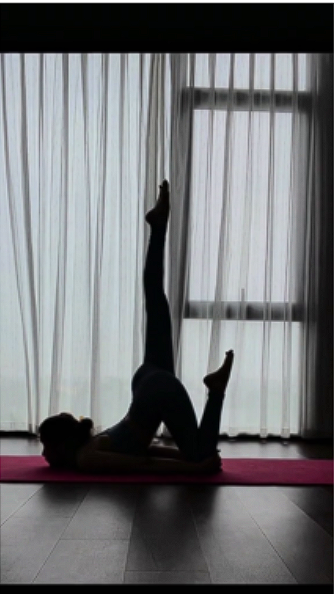}
\includegraphics[width=0.143\textwidth]{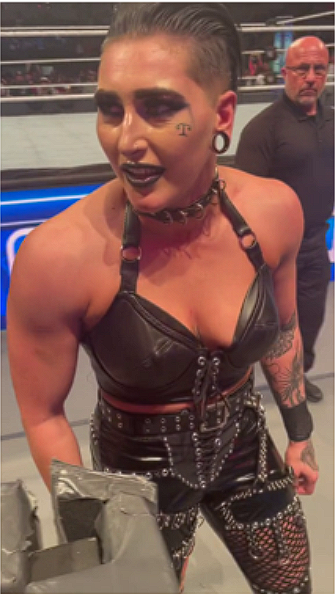}
\includegraphics[width=0.143\textwidth]{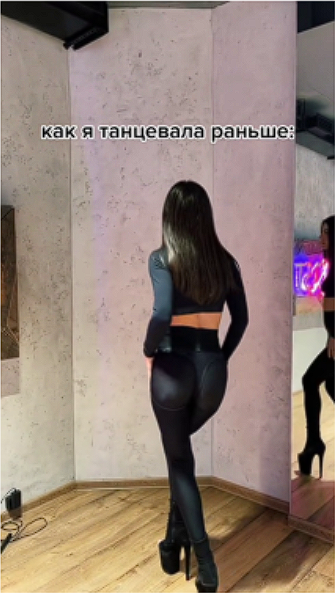}
\caption{Examples of misclassification cases where predicted labels differ from ground truth. From left to right, Safe as Harmful Content, Harmful Content as Adult Content and Adult Content as Harmful Content.}
\label{fig:misclass_cases}
\end{figure}
\begin{figure}[h]
\centering
\includegraphics[width=0.143\textwidth]{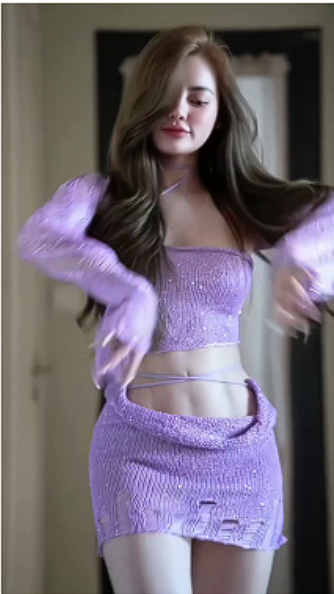}
\includegraphics[width=0.18\textwidth]{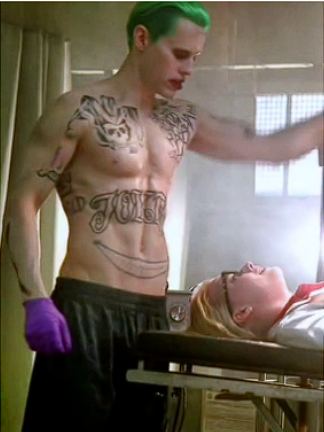}
\includegraphics[width=0.143\textwidth]{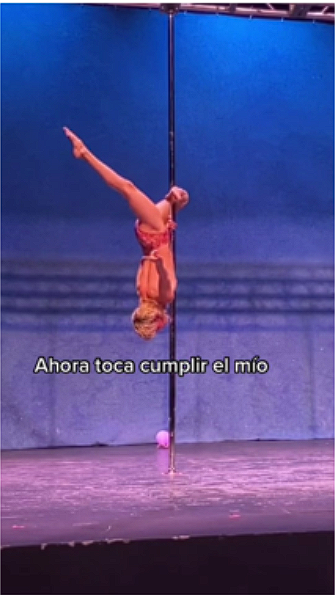}
\caption{Examples of overlapping-label cases where videos could plausibly belong to multiple categories. From left to right, Harmful Content also Adult Content, Suicide also Adult Content and Adult Content also Harmful Content.}
\label{fig:overlap_cases}
\end{figure}
We observed two main error types. The first is misclassification, where the predicted label differs from the ground truth (Figure~\ref{fig:misclass_cases}). For example, \textit{Safe} videos predicted as \textit{Harmful Content} due to poses resembling unsafe contexts, \textit{Harmful Content} misclassified as \textit{Adult Content} because of clothing or makeup cues, and \textit{Adult Content} predicted as \textit{Harmful Content} due to aggressive thematic elements. In some cases, these predictions may still be justifiable under stricter moderation perspectives. The second type is overlapping-label cases, where videos plausibly belong to more than one harmful category (Figure~\ref{fig:overlap_cases}), such as content mixing aggressive and sexual elements, suicide-related context with adult cues, or mature visuals with harmful implications. These findings suggest that future work should explore multi-label classification and improved multimodal fusion to address ambiguous cases.
\subsection{Noisy Cases}
\label{sec:noisy_cases}
\begin{figure}[h]
\centering
\subfloat[]{\includegraphics[width=0.143\textwidth]{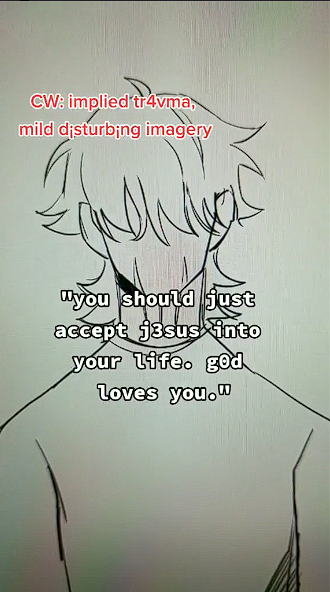}}
\hspace{2mm}
\subfloat[]{\includegraphics[width=0.143\textwidth]{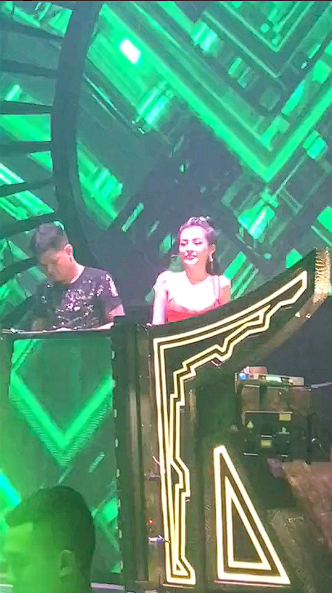}}
\hspace{2mm}
\subfloat[]{\includegraphics[width=0.15\textwidth]{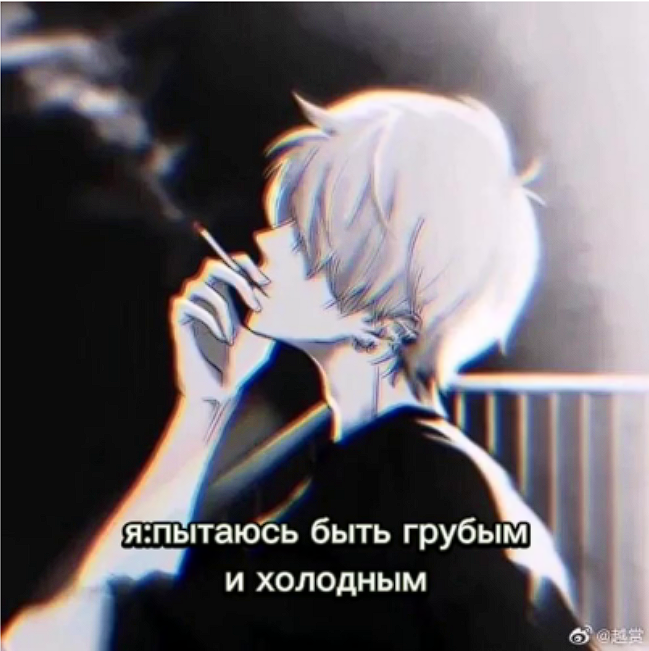}}
\caption{Examples of noisy cases. From left to right: 
(a) stylized and numeric text with blurred, flickering frames, 
(b) nightclub environment with light and audio noise, 
(c) code-switched content with unrelated audio.}
\label{fig:noisy_cases}
\end{figure}
As shown in Figure~\ref{fig:noisy_cases}, the case (a) combines multiple types of noise: teencode in the opening frames, numeric overlays in later segments, loud rock-style background audio, and flickering visuals. The case (b) suffers from strong lighting distortions and audio that lacks linguistic content, while (c) mixes cross-lingual subtitles with English audio unrelated to the annotated label. Despite these imperfections, MTikGuard successfully classified all cases into their correct categories - (a) \textit{Suicide}, (b) \textit{Adult Content}, (c) \textit{Suicide}, confirming that the multimodal fusion design provides robustness even under noisy and ambiguous conditions typical of TikTok videos.

\end{document}